\documentclass[sigconf]{acmart}
\AtBeginDocument{%
  }

\setcopyright{cc}
\copyrightyear{2026}
\acmYear{2026}
\setcctype{by}
\acmConference[SIGIR '26] {Proceedings of the 49th International ACM SIGIR Conference on Research and Development in Information Retrieval}{July 20--24, 2026}{Melbourne, VIC, Australia.}
\acmBooktitle{Proceedings of the 49th International ACM SIGIR Conference on Research and Development in Information Retrieval (SIGIR '26), July 20--24, 2026, Melbourne, VIC, Australia}
\acmDOI{10.1145/3805712.3808366}
\acmISBN{979-8-4007-2599-9/2026/07}

\usepackage{xcolor}
\usepackage{listings}
\lstdefinestyle{promptblockstyle}{
  basicstyle=\ttfamily\footnotesize,
  breaklines=true,
  columns=fullflexible,
  keepspaces=true,
  frame=single,
  framerule=0.3pt,
  rulecolor=\color{black!25},
  backgroundcolor=\color{gray!3},
  xleftmargin=4pt,
  xrightmargin=4pt,
  aboveskip=4pt,
  belowskip=4pt
}
\usepackage[inline]{enumitem}
\usepackage{pifont}
\newcommand{\cmark}{\ding{51}}%
\newcommand{\xmark}{\ding{55}}%
\usepackage{tikz}
\usetikzlibrary{positioning,arrows.meta,fit,shapes.misc}

\newcommand{\system}{\textsc{MuseKG}}

\usepackage{hyperref}

\graphicspath{{figures/}{samples/figures/}}
\begin{document}

\title{\system: An Interactive Knowledge Graph Over Museum Collections}

\author{Jinhao Li}
\orcid{0009-0006-9301-5579}
\affiliation{%
  \institution{The University of Melbourne}
  \department{School of Computing and Information Systems}
  \city{Melbourne}
  \state{Victoria}
  \country{Australia}
}
\email{jinhao3@student.unimelb.edu.au}

\author{Jianzhong Qi}
\orcid{0000-0001-6501-9050}
\affiliation{%
  \institution{The University of Melbourne}
  \department{School of Computing and Information Systems}
  \city{Melbourne}
  \state{Victoria}
  \country{Australia}
}
\email{jianzhong.qi@unimelb.edu.au}

\author{Soyeon~Caren~Han}
\orcid{0000-0002-1948-6819}
\affiliation{%
  \institution{The University of Melbourne}
  \department{School of Computing and Information Systems}
  \city{Melbourne}
  \state{Victoria}
  \country{Australia}
}
\email{caren.han@unimelb.edu.au}

\author{Eun-Jung~Holden}
\orcid{0000-0002-8752-1639}
\affiliation{%
  \institution{The University of Melbourne}
  \department{School of Computing and Information Systems}
  \city{Melbourne}
  \state{Victoria}
  \country{Australia}
}
\email{eunjung.holden@unimelb.edu.au}

\renewcommand{\shortauthors}{Jinhao Li, Jianzhong Qi, Soyeon Caren Han, and Eun-Jung Holden}

\begin{abstract}
Digitisation in the cultural heritage sector has produced large but fragmented repositories of museum collection data, spanning structured catalogue records, images, and unstructured descriptions. Existing museum information systems often make it difficult to integrate these sources into a unified, queryable representation that supports relation-aware exploration.
We present \textbf{MuseKG}, an interactive knowledge graph system that organises heterogeneous museum data into a typed graph that links objects, people, organisations, images, image-derived labels, and extracted semantic entities within a coherent schema. MuseKG supports natural-language queries by grounding user questions to graph entities and retrieving a compact neighbourhood of evidence for answer generation.
Through an interactive demonstration on real museum collections, we show that MuseKG supports common exploration tasks such as attribute lookup, relation exploration, and relation-aware retrieval, with answers that remain inspectable via explicit graph structures.
\end{abstract}




\maketitle

\section{Introduction}
Digitisation efforts across the cultural heritage sector have created vast yet fragmented repositories of artefact information, spanning text-based catalogues, image archives, and multimedia documentation. Despite this abundance, museum collections remain difficult to search, connect, and interpret holistically. The key obstacle is not only scale, but \emph{semantic disconnection}: object descriptions, relationships, and contextual metadata are often distributed across heterogeneous systems, which limits cross-collection discovery and interpretive reasoning.

Recent advances in large language models (LLMs)~\citep{agarwal2025gpt, team2025gemma} and knowledge graphs (KGs)~\citep{peng2023knowledge} have renewed interest in semantically enriched museum infrastructures, where LLMs provide flexible natural-language access and KGs provide explicit structure for relation-aware retrieval and interpretable evidence. 
In cultural heritage, this direction is supported by large-scale knowledge graph efforts that publish interoperable, queryable cultural heritage data (e.g., ArCo~\citep{carriero2019arco}, and InTaVia~\citep{schlogl2025intavia}). Complementing these resources, museum-oriented pipelines have explored practical KG construction from collection catalogues and linking to external hubs such as Wikidata, exemplified by Heritage Connector~\citep{dutia2021heritage}. 
Beyond linking, recent work has examined the complementary roles of knowledge graphs and large language models in virtual museums
settings~\cite{vasic2025knowledge}, and applied knowledge graphs with deep learning to automate cultural heritage management~\cite{huang2023using}.
At the interaction layer, recent work has investigated translating natural-language questions into executable SPARQL queries over CIDOC-CRM knowledge bases~\citep{varagnolo2025translating} and using LLMs to generate CIDOC-CRM SPARQL queries directly~\citep{mountantonakis2025generating}, with broader NL2SPARQL techniques such as chain-of-thought prompting further improving query generation in general KG settings~\citep{zahera2024generating, avila2024experiments, d2025investigating}. \emph{Together, these efforts indicate a broader shift toward intelligent, multimodal, and semantically integrated cultural heritage retrieval and access}.

However, existing approaches often cover only part of the end-to-end pipeline, leaving a gap for an \emph{interactive} system that connects heterogeneous collection data, KG construction, and natural-language querying. Standards- and KG-centric efforts typically assume expert SPARQL use and substantial modelling~\citep{carriero2019arco, schlogl2025intavia}, while automated KG construction methods~\citep{guo2022automatic,
gohsen2024assisted} and catalogue-to-KG pipelines emphasise linking/enrichment but provide limited explainable query experiences for end users~\citep{dutia2021heritage}. NL2SPARQL and LLM-based query generation improve accessibility yet can be brittle under complex schemas, motivating KG-grounded execution with inspectable evidence~\citep{varagnolo2025translating, mountantonakis2025generating, zahera2024generating}. Although interactive KG navigation tools show the value of neighbourhood-based inspection~\citep{wang2023kgnav}, they are rarely integrated with museum schemas and multimodal artefacts; consequently, museum retrieval still largely relies on keyword or field-based search, which struggles with relation traversal and evidence-grounded answers.

In this paper, we present \underline{Muse}um \underline{K}nowledge \underline{G}raph (\textbf{\system}), an interactive KG system for museum collections that integrates structured records with media-derived labels in a typed property graph (cf.~Figure~\ref{fig:example_kg}). \system\ links objects, people, organisations, images, image-derived labels, and extracted semantic entities into a coherent schema and exposes a lightweight natural-language interface that maps user questions to executable KG operations for interpretable retrieval over attributes and relations. We demonstrate \system\ through interactive scenarios including (i) attribute lookup for artefact inspection, (ii) relation exploration for provenance-style discovery, and (iii) relation-aware queries that retrieve attributes of related entities. To support the demonstration, we construct a small benchmark via KG-grounded and LLM-guided question--answer generation, avoiding reliance on manually authored query templates. Our contributions are:

(1)~MuseKG, an interactive KG system for constructing and querying museum collections via natural language.

(2)~A lightweight KG construction pipeline and NL query interface for attribute lookup, relation exploration, and relation-aware retrieval.

(3)~Demonstration scenarios on a real museum collection showing interpretable, evidence-grounded answers.

\begin{figure*}[t]
  \includegraphics[width=\textwidth]{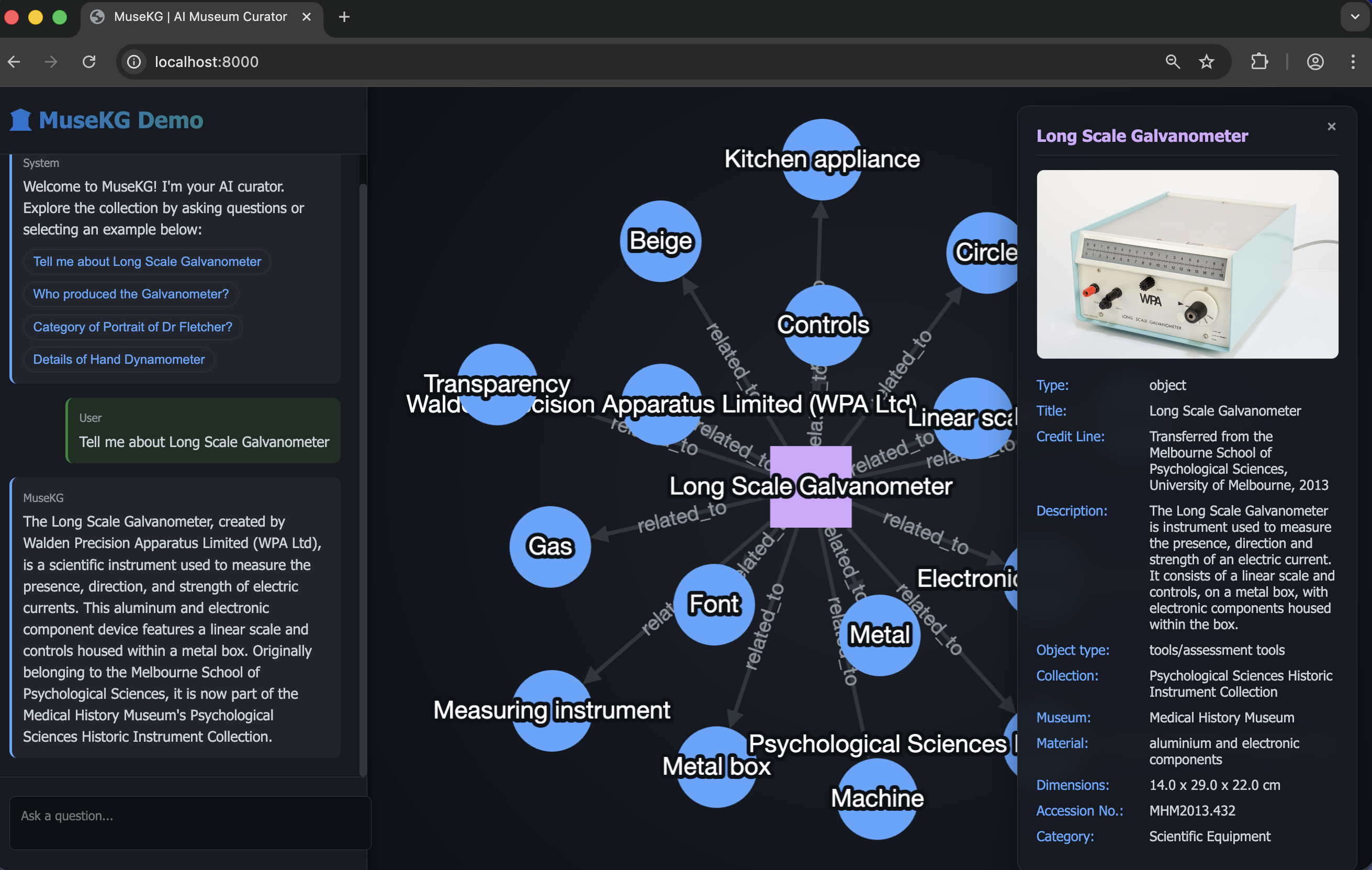}
  \caption{MuseKG interactive demo interface. Left: chat-style natural-language query panel with example prompts. Center: visualisation of the retrieved KG neighbourhood around the selected artefact. Right: entity detail card showing the artefact image and key metadata fields grounded in the KG.}
  \Description{Screenshot of the MuseKG demo web interface. A left sidebar shows a chat panel with example questions and a conversation about the ``Long Scale Galvanometer''. The center area displays a node-link graph with the selected artefact in the middle connected to related nodes such as labels and entities. A right-hand panel shows the artefact photo and structured metadata including title, description, collection, material, dimensions, and accession number.}
  \Description{Screenshot of the MuseKG demo interface with a chat-style query panel on the left, a graph neighbourhood visualization in the center, and an entity detail panel on the right.}
  \label{fig:teaser}
\end{figure*}

\noindent

\section{System Components}\label{sec:method}

\system\ harvests information about museum collections, organises it into a KG, and provides a natural-language interface for users to query and explore the data.

We define the KG as a typed property graph
$G = (V, E, \tau, \rho, A)$,
where $V$ is a finite set of nodes representing entities in the museum collections.
Let $E$ be a finite set of directed edges, and let $\mathrm{src}, \mathrm{dst}: E \to V$
map each edge to its source and destination nodes.
The function $\tau: V \to \mathcal{T}$ assigns each node a type from a finite set of permissible node types $\mathcal{T}$ (e.g., \textit{object}, \textit{person}, \textit{organisation}).
The function $\rho: E \to \mathcal{R}$ assigns each edge a relation label from a finite set of relations
$\mathcal{R}$; these labels are normalised via a mapping that converts raw relationship identifiers in the museum collection data into canonical relations.
Finally, $A: V \to (\mathcal{K} \to \mathcal{U})$ maps each node to an attribute dictionary from keys
$\mathcal{K}$ to attribute values in $\mathcal{U}$.

\begin{figure}[t]
    \centering
    \resizebox{0.55\columnwidth}{!}{%
    \begin{tikzpicture}[
        font=\small,
        bigbox/.style={
            draw,
            rounded corners,
            align=left,
            inner sep=6pt,
            minimum width=5cm,
            fill=gray!5
        },
        kgbox/.style={
            draw,
            rounded corners,
            align=center,
            inner sep=6pt,
            minimum width=4.5cm
        },
        arrow/.style={
            -{Latex[length=3mm]},
            thick
        }
    ]

    \node[bigbox] (m1) {
        \textbf{Module 1: KG constructor}\\[2pt]
        \footnotesize
        Inputs: Raw museum collection records (JSON).\\
        \footnotesize
        Steps: Normalisation, entity identification,\\
        \footnotesize
        node \& edge creation, deduplication, and schema checks.
    };

    \node[kgbox, below=7mm of m1] (kg) {
        \textbf{Museum collections KG}\\[2pt]
        \footnotesize Typed property graph (\system) over\\
        \footnotesize objects, people, organisations, images, image-derived labels,\\
        \footnotesize and extracted semantic entities.
    };

    \node[bigbox, below=7mm of kg] (m2) {
        \textbf{Module 2: NL query interface}\\[2pt]
        \footnotesize
        Inputs: User query in natural language.\\
        \footnotesize
        Steps: LLM-based entity extraction, KG context\\
        \footnotesize
        retrieval (attributes \& neighbours), and answer generation.\\
        \footnotesize
        Output: Natural-language answer grounded in \system.
    };

    \draw[arrow] (m1) -- node[right]{builds} (kg);
    \draw[arrow] (kg) -- node[right]{provides KG context to} (m2);

    \end{tikzpicture}%
    }
    \caption{System overview of \system. Module 1 constructs the
    museum collections KG (\system) from records.
    Module 2 takes a user query, retrieves KG context, and uses an
    LLM to generate a natural-language answer grounded in \system.}
    \Description{System overview showing raw museum collection records passed to a knowledge graph constructor, producing a typed museum collections graph that provides context to a natural-language query interface.}
    \label{fig:system-overview}
\end{figure}

As shown in \figureautorefname~\ref{fig:system-overview}, \system\ consists of two core components: (i)~a KG constructor and (ii)~a natural language query interface. We detail these components below.

\subsection{Knowledge Graph Constructor}\label{subsec:kg_constructor}

We start with the source data and explain how \system\ constructs a KG from such data. 

\textbf{Source Data.} \system\ is built upon a museum collection dataset $\mathcal{D}$ composed of collections owned by The University of Melbourne, in the form of JSON records. Each record in $\mathcal{D}$ contains: (i) object metadata (e.g., title, material description, accession number, and various dates such as production and acquisition dates) and (ii) relational fields that reference people, organisations, related objects, and media-derived labels (e.g., image labels from object photographs). Our methodology can be easily \emph{extended} to datasets following similar formats.
Table \ref{tab:dataset_stats} summarises dataset statistics.

\begin{table}[t]
  \centering
  \caption{Dataset Statistics}
  \label{tab:dataset_stats}
  \small
  \setlength{\tabcolsep}{6pt}
  \renewcommand{\arraystretch}{0.9}
  \setlength{\cmidrulekern}{2pt}
  \setlength{\cmidrulesep}{2pt}
  \begin{tabular}{@{}l r  l r@{}}
    \toprule
    \textbf{Name} & \textbf{Count} & \textbf{Name} & \textbf{Count} \\
    \midrule
    \multicolumn{4}{@{}l}{\textbf{Overview}}\\[-2pt]
    Unique Object IDs     & 15{,}829 & Unique Collection IDs & 3 \\[2pt]

    \multicolumn{2}{@{}l}{\textbf{Object Attributes}} & \multicolumn{2}{l@{}}{\textbf{}}\\[-2pt]
    Name (Title)          & 15{,}828 & Credit Line          & 7{,}732 \\
    Material Description  & 13{,}598 & Production Date      & 12{,}917 \\
    Description           & 15{,}082 & Object Type          & 15{,}734 \\
    Accession Number      & 15{,}825 & History Category     & 10{,}378 \\
    Measurements          & 13{,}080 &          &  \\[2pt]

    \multicolumn{4}{@{}l}{\textbf{Relationships}}\\[-2pt]
    Primary Producer (Person) & 8{,}367 & Related Object & 2{,}967 \\
    Images              & 45{,}713 & & \\
    \bottomrule
  \end{tabular}
\end{table}

\begin{figure}[t]
    \centering
    \includegraphics[width=\linewidth]{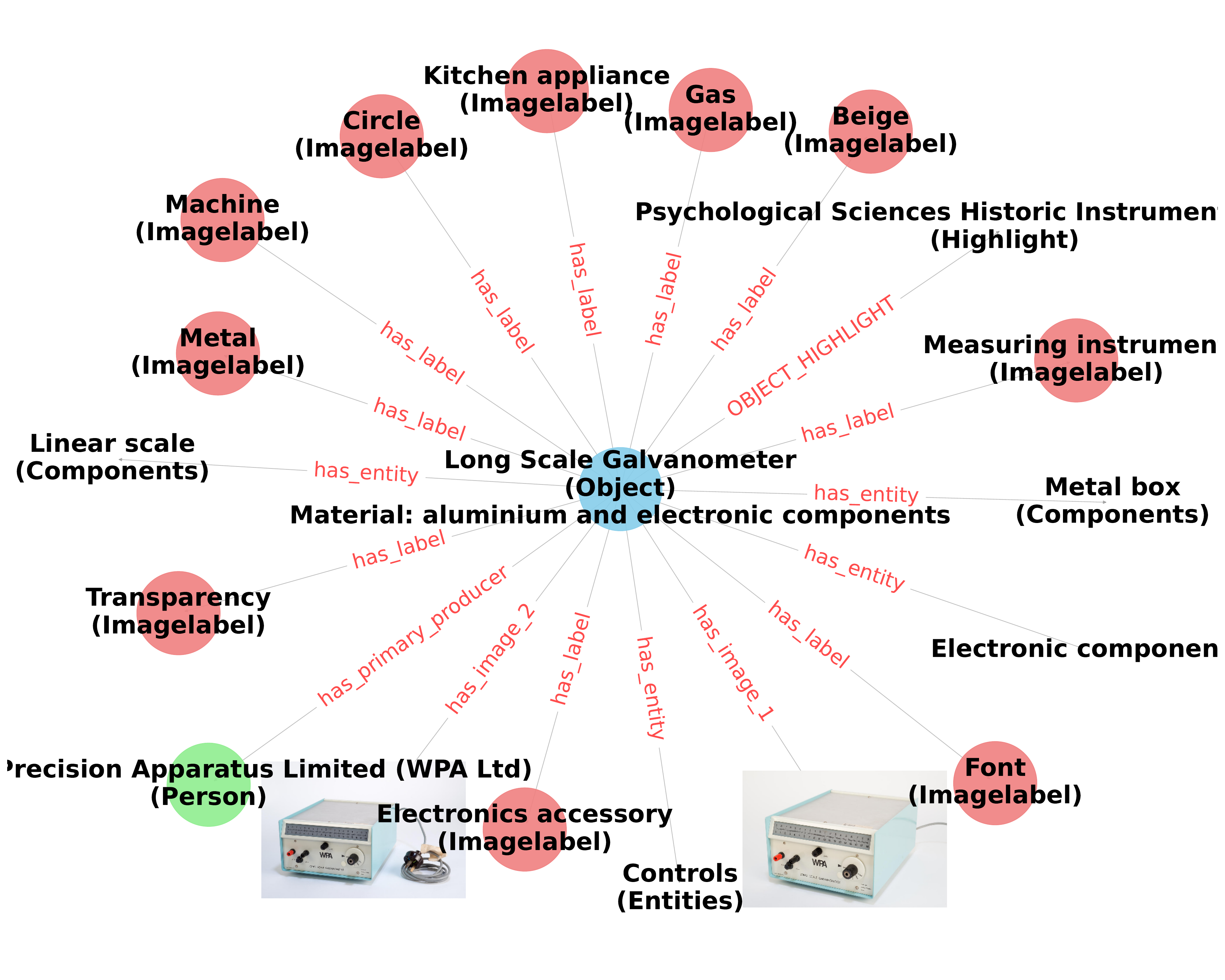}
    \caption{Visualisation of an example KG subgraph constructed from a single record. The central object node and its neighbouring nodes (image labels, components, entities, and people) illustrate the attributes that are referenced in the subsequent query example.}
    \Description{Example knowledge graph subgraph centred on the Long Scale Galvanometer object, connected to image labels, components, semantic entities, images, and a primary producer node.}
    \label{fig:example_kg}
\end{figure}

Below is an example record representing an object \texttt{OBJ123}, focusing on the data fields considered by \system. This record details the object's name, material description, primary producer, and associated image.
\begin{lstlisting}[style=promptblockstyle]
{
  "opacObjectId": "OBJ123",
  "name": "Long Scale Galvanometer",
  "material_desc": "aluminium and electronic components",
  "relationships": [{
    "type": "object_prod_pri_person",
    "title": "Walden Precision Apparatus Limited"
  }],
  "images": [{"imageId": "20208"}]
}
\end{lstlisting}

\textbf{KG Constructor.}
Our KG constructor maps source data records to graph nodes and edges through the following steps: 

(1)~\textbf{Normalisation:} Textual fields are lowercased, whitespace collapsed, and punctuation/quotes stripped to stabilise matching during entity resolution and querying.
    
(2)~\textbf{Entity Identification and Node Creation:} For each distinct entity identified from the source data, a corresponding node is created in the KG. An ``entity identifier'' in \system\ refers to the unique value used to represent a distinct entity as a node. 
    Nodes are created for the following entities: (i)~each \texttt{opacObjectId} from the main object records; (ii)~entities referenced in \texttt{relationshipsCollection} (e.g., persons, organisations, or related objects), using their provided Id (\texttt{relatedRecordId}) as identifiers; (iii)~imageId extracted from \texttt{imagesCollection}; and (iv)~named entities (e.g., places, persons, concepts) detected and linked from textual fields (e.g., descriptions).

(3)~\textbf{Node Type Assignment:} The node type mapping $\tau$ assigns a specific type to each created node: 
    \texttt{object} for nodes created from \texttt{opacObjectId}; 
    \texttt{person} or \texttt{organisation} for nodes constructed from objects in the \texttt{relationshipsCollection}, as indicated by their \texttt{relatedRecordType} field;
    \texttt{image} for nodes created from \texttt{imageId}; image label for media-derived labels; and semantic entity types such as place, person, or concept for entities extracted from textual fields.

(4)~\textbf{Relation Labelling and Edge Creation:} Relationship identifiers (e.g., \texttt{object\_prod\_pri\_person}) are mapped to a fixed vocabulary $\mathcal{R}$ (for simplicity). Edges are then created between the corresponding nodes based on these normalised relation labels.
    
(5)~\textbf{Deduplication:} Nodes are initially created for each unique entity identifier. Duplicate nodes, identified by canonical titles or accession numbers, are then merged, retaining the first-seen attribute payload. Duplicate edges are detected and removed based on their
$(\mathrm{src}(e), \rho(e), \mathrm{dst}(e))$ triples.
    
(6)~\textbf{Schema Validation:} Extracted attributes for each node are restricted to a predefined schema $\mathcal{K}$. Other keys are dropped. 

The complete set $\mathcal{R}$ comprises seven relation types
grouped into four categories: \emph{provenance}
(primary/secondary producer and organisation links),
\emph{inter-object} (related object),
\emph{visual} (image-derived labels), and
\emph{semantic} (named entities extracted from descriptions).
These are normalised from the source museum system's raw
identifiers (e.g., \texttt{object\_prod\_pri\_person}
$\to$ \texttt{has\_primary\_\allowbreak producer}) rather than a
standard ontology such as CIDOC-CRM.
The resulting KG contains 23{,}263 nodes and 75{,}375 edges
constructed from 15{,}829 object records across 3~collections.

\textbf{Running Example: From Record to Graph.}
Consider the example JSON record for object \texttt{OBJ123} (from the ``Source Data''  paragraph). 
Our KG constructor processes this record to generate the nodes and edges as visualised in \figureautorefname~\ref{fig:example_kg}. The figure illustrates the central object node (`Long Scale Galvanometer') with its attributes (e.g., title, material description) and its direct connections to other nodes, such as its primary producer (`Walden Precision Apparatus Limited (WPA Ltd)'), and associated images.

\subsection{Natural Language Query Interface}\label{subsec:query_interface}
Our \system\ query interface  answers NL queries by Retrieval-Augmented Generation (RAG)~\citep{lewis2020retrieval} over the KG in two main stages:

\textbf{Context Retrieval from KG.}
Upon receiving a query, \system\ first identifies key entities in the question. An LLM is employed to extract key entities (e.g., an object title) from the NL question. For each identified entity, \system\ then retrieves a textual context from the KG. This context is formed by all attributes associated with the entity's node and its relations to one-hop neighbours. This step effectively grounds the query within the rich, interconnected data of the KG, providing relevant factual evidence. Note that \system\ can be extended to multi-hop retrieval to answer more complex queries.

For example, given a query ``What are the measurements for the Long Scale Galvanometer?", \system\ first identifies `Long Scale Galvanometer' as the key entity. It then retrieves its attributes (including `measurements') and any directly connected entities. This collected information forms the context:
\begin{lstlisting}[style=promptblockstyle]
Object: Long Scale Galvanometer
- material_desc: aluminium and electronic components
- measurements: 14.0 x 29.0 x 22.0 cm
- accession_no: MHM2013.432
- credit_line: Transferred from the Melbourne School of Psychological Sciences, University of Melbourne, 2013
... (truncated for brevity)
\end{lstlisting}

\textbf{LLM-Based Answer Generation.}
The retrieved textual context, along with the original NL query, is then fed into an LLM, which is prompted to synthesise a coherent answer, as shown below:
\begin{lstlisting}[style=promptblockstyle]
Answer using ONLY the KG context. Return ONLY the final answer.
Context: {context}
Question: {question}
Answer:
\end{lstlisting}

This RAG-based approach allows \system\ to combine the LLM's NL  understanding and generation capabilities with the structured knowledge derived from the KG, thereby mitigating hallucinations often observed in LLM-only systems. 

\section{System Demonstration}
\label{sec:demo}

We demonstrate \system\ as an interactive museum exploration and question-answering system that combines \emph{executable} knowledge-graph retrieval with \emph{evidence-grounded} natural-language responses. Figure~\ref{fig:teaser} shows the end-to-end demo interface: a user can type a free-form question (or click a suggested query), \system\ resolves the main entity mention(s), retrieves a compact KG neighbourhood as evidence, visualises the retrieved subgraph for inspection, and produces a concise answer that is grounded in the retrieved KG facts. The key design goal of the demo is to make the system \emph{usable and inspectable}: users can see \emph{what} information was retrieved (entity attributes and neighbours), \emph{where} it came from in the graph, and \emph{how} the answer is supported by explicit relations and metadata.

\subsection{Demo Tasks}
The demonstration centres on three common interaction patterns in museum search that map directly to \system's KG operations. For \textbf{object inspection (attribute lookup)}, users ask for specific fields (e.g., materials, dimensions, accession number), and \system\ grounds the query to the object node and surfaces the retrieved values in the entity panel (Figure~\ref{fig:teaser}, right). For \textbf{provenance and connections (relation exploration)}, users follow relations such as producers or related objects, with connected entities shown in the graph neighbourhood view and the entity panel (Figure~\ref{fig:teaser}, center/right). For \textbf{explainable relation-aware queries (relation $\rightarrow$ attribute)}, \system\ traverses the required edge(s) and retrieves the target attribute, highlighting the traversed path in the graph view to provide provenance for the answer.

\subsection{User Interface and Interaction Workflow}
As illustrated in Figure~\ref{fig:teaser}, the demo presents a full query-to-answer loop in a single view for fast, inspection-driven interaction. Users type a natural-language question or select an example prompt (left), after which \system\ grounds the query to key entity mention(s) (typically an object title) and retrieves KG evidence consisting of schema-constrained attributes and one-hop neighbours connected by typed relations. The interface then focuses the graph neighbourhood view on the retrieved subgraph (center) to enable interactive verification and exploration, while an entity panel (right) surfaces structured metadata and associated images for any selected node. Finally, an LLM generates a concise answer using only the retrieved KG evidence, aligning the response with the visible attributes and relations so users can easily cross-check provenance. The demo system, source code, and supplementary materials are available at:
\url{https://github.com/JinhaoLee/MuseKG}.

\subsection{Demonstration Scenarios}
\label{subsec:demo_scenarios}
We demonstrate \system\ with five representative interactions: (1) an entity-grounded provenance query such as \textit{``Who produced the Long Scale Galvanometer?''}, where \system\ grounds the object mention and follows \texttt{has\_primary\_producer} to retrieve and visualise the linked producer; (2) attribute lookup for object inspection (e.g., dimensions or accession number), returning exact catalog fields surfaced in the entity panel (Figure~\ref{fig:teaser}); (3) multimodal exploration via clicking the graph neighbourhood to pivot across linked entities (e.g., image labels or concepts) while inspecting high-resolution images and metadata in the panel; (4) robustness to noisy input (e.g., \textit{``glvanometer''}) via lightweight entity resolution before KG-grounded retrieval; and (5) relation-aware retrieval (relation $\rightarrow$ attribute) such as \textit{``What is the accession number of the entity associated with the Certificate of Passing First Year of Bachelor of Laws?''}, where \system\ traverses \texttt{has\_entity} and returns the requested attribute with the traversed path visible in the graph view.

\begin{table}[t]
\centering
\caption{Evaluation accuracy across three query categories. ``Cat. prompt'' indicates whether the method uses category-specific prompt engineering.}
\label{tab:eval}
\label{tab:evaluation-results}
\small
\begin{tabular}{lcccc}
\toprule
Pipeline & C1 & C2 & C3 & Cat. prompt \\
\midrule
\multicolumn{5}{@{}l}{\textbf{Gemma-3-4B}}\\[-2pt]
\texttt{LLM Zero-shot}      & 0.20 & 0.04 & 0.02 & \xmark \\
\texttt{LLM Few-shot}       & 0.50 & 0.04 & 0.06 & \cmark \\
\texttt{LLM SPARQL-prompt}  & 0.80 & 0.32 & 0.12 & \cmark \\
\texttt{MuseKG (Ours)}      & \textbf{0.84} & \textbf{0.42} & \textbf{0.34} & \xmark \\
\addlinespace[2pt]
\multicolumn{5}{@{}l}{\textbf{Gemma-3-12B}}\\[-2pt]
\texttt{LLM Zero-shot}      & 0.26 & 0.04 & 0.02 & \xmark \\
\texttt{LLM Few-shot}       & 0.52 & 0.06 & 0.10 & \cmark \\
\texttt{LLM SPARQL-prompt}  & 0.76 & 0.28 & 0.10 & \cmark \\
\texttt{MuseKG (Ours)}      & \textbf{0.84} & \textbf{0.42} & \textbf{0.36} & \xmark \\
\multicolumn{5}{@{}l}{\textbf{gpt-oss-20b}}\\[-2pt]
\texttt{LLM Zero-shot}      & 0.18 & 0.04 & 0.10 & \xmark \\
\texttt{LLM Few-shot}       & 0.38 & 0.10 & 0.08 & \cmark \\
\texttt{LLM SPARQL-prompt}  & 0.72 & 0.22 & 0.10 & \cmark \\
\texttt{MuseKG (Ours)}      & \textbf{0.84} & \textbf{0.50} & \textbf{0.32} & \xmark \\
\bottomrule
\end{tabular}
\end{table}

\section{System Validation}

\textbf{Benchmark Construction.}
We construct a benchmark of 150 questions (50 per category) from
the source museum JSON records. For each sampled object, we
prompt \texttt{Gemini-2.5-flash-lite} with a category-specific
instruction that presents the raw JSON and asks for a single
question--answer pair together with a structured KG query
description. C1 prompts target attribute-lookup questions
(e.g., retrieving an object's measurements); C2 prompts target
relation-neighbour questions using an explicit relationship from
the KG schema; C3 prompts require first traversing a relation and
then retrieving an attribute of the target node. We parse the
output as JSON and discard malformed responses.
Ground-truth answers are derived directly from the KG, and
correctness is assessed by an independent LLM judge
(Gemma-3-12B) that compares each system output against the
expected answer.

Table~\ref{tab:evaluation-results} summarises a supporting validation of \system\ across three backbone LLMs. We consider three query types that align with the demo scenarios: \textbf{(C1) attribute lookup} for object inspection (e.g., measurements, materials), \textbf{(C2) relation exploration} to retrieve linked entities (e.g., producers or related objects), and \textbf{(C3) relation-aware attribute retrieval} that follows a relation and then returns an attribute of the connected entity.
We compare \system\ with LLM-only prompting baselines, including zero-shot and few-shot answering, as well as an NL2SPARQL prompting pipeline that generates an executable query over the KG. While the few-shot and NL2SPARQL baselines require category-specific prompt tuning, \system\ uses a single generic KG-RAG prompt and consistently achieves the strongest accuracy overall. The NL2SPARQL baseline is the most competitive among the baselines, indicating that exposing KG structure is helpful; nevertheless, it remains weaker on relation-centric queries (C2 and C3), where \system's explicit KG-grounded execution provides the largest gains.

\section{Conclusion}
We presented \system, an interactive system that bridges the gap between fragmented museum records and accessible natural language querying. Our demonstration highlights how \system\ combines the structural precision of a knowledge graph with the flexibility of LLMs to provide inspectable, evidence-grounded answers. Evaluation on our benchmarks confirms that \system\ consistently outperforms baselines on relation-centric tasks. 
Currently, MuseKG resolves a single main entity per query, which limits its ability to answer comparative questions spanning multiple entities or open-ended questions that reference no specific object.
Future work will focus on scaling the pipeline to cross-collection discovery and evolving \system\ into an \emph{agent-based framework}, capable of autonomously planning and executing complex investigations across heterogeneous museum networks.

\section*{Acknowledgments}
We thank the University of Melbourne's MDHS Museums for providing access to the collections and data that enabled this study.
This research was supported by The University of Melbourne’s Research Computing Services and the Petascale Campus Initiative.
Jinhao Li is supported by the Melbourne Research Scholarship.
Jianzhong Qi is in part supported by the Australian Research Council (ARC) via Discovery Project DP240101006 and Future Fellowship FT240100170.










\bibliographystyle{ACM-Reference-Format}
\bibliography{sample-base}


\end{document}